\documentclass[11pt,a4paper]{article}

\usepackage[hyperref]{acl2019}
\usepackage{times}
\usepackage{latexsym}
\usepackage{linguex} 
\usepackage{tikz}
\usepackage{pgf}
\usetikzlibrary{shapes,external}
\usetikzlibrary{shapes.multipart}
\usetikzlibrary{calc}

\usepackage{amsmath}
\usepackage{amssymb}
\usepackage{mathtools}
\usepackage{proof}

\newlength{\arrow}
\newcommand*{\myrightarrow}[1]{\xrightarrow{\mathmakebox[\arrow]{\text{\scriptsize #1}}}}

\usepackage{color}
\definecolor{first}{RGB}{82,82,82}
\definecolor{enc}{RGB}{253,192,134}
\definecolor{dec}{RGB}{127,201,127}
\definecolor{dec2}{RGB}{140,220,140}
\definecolor{emb}{RGB}{190,174,212}

\usepackage{float}

\usepackage{subcaption}

\usepackage{tabularx}
\usepackage{booktabs}
\newcolumntype{m}{>{\hsize=.65\hsize}X}
\newcolumntype{s}{>{\hsize=.4\hsize}c}
\newcolumntype{u}{>{\hsize=.25\hsize}X}
\usepackage{multicol}
\usepackage{multirow}

\usepackage{url}

\aclfinalcopy 
\def\aclpaperid{26} 

\title{Constructive Type-Logical Supertagging with Self-Attention Networks}

\author{Konstantinos Kogkalidis \\
  Utrecht University\\
  \texttt{k.kogkalidis@uu.nl} \\\And
  Michael Moortgat\\
  Utrecht University\\
  \texttt{m.j.moortgat@uu.nl} \\\And
  Tejaswini Deoskar \\
  Utrecht University\\
  \texttt{t.deoskar@uu.nl} \\
  }

\date{}

\begin{document}
\maketitle
\begin{abstract}
We propose a novel application of self-attention networks towards grammar induction. 
We present an attention-based supertagger for a refined type-logical grammar, trained on constructing types inductively.
In addition to achieving a high overall type accuracy, our model is able to learn the syntax of the grammar's type system along with its denotational semantics. 
This lifts the closed world assumption commonly made by lexicalized grammar supertaggers, greatly enhancing its generalization potential. 
This is evidenced both by its adequate accuracy over sparse word types and its ability to correctly construct complex types never seen during training, which, to the best of our knowledge, was as of yet unaccomplished.
\end{abstract}

\section{Introduction}
Categorial Grammars, in their various incarnations, posit a functional view on parsing: words are assigned simple or complex categories (or: types); their composition is modeled in terms of functor-argument relationships. Complex categories wear their combinatorics on their sleeve, which means that most of the phrasal structure is internalized within the categories themselves; performing the categorial assignment process for a sequence of words, i.e. \textit{supertagging}, amounts to almost parsing~\cite{supertagging}.

In machine learning literature, supertagging is commonly viewed as a particular case of sequence labeling~\cite{Graves2012}.
This perspective points to the immediate applicability of established, high-performing neural architectures; indeed, recurrent models have successfully been employed (e.g. within the context of Combinatory Categorial Grammars (CCG)~\cite{steedman00}), achieving impressive results~\cite{vaswani2016lstm}.
However, this perspective comes at a cost; the supertagger's co-domain, i.e., the different categories it may assign, is considered fixed, as defined by the set of unique categories in the training data.
Additionally, some categories have disproportionately low frequencies compared to the more common ones, leading to severe sparsity issues. 
Since under-represented categories are very hard to learn, in practice models are evaluated and compared based on their accuracy over categories with occurrence counts above a certain threshold, a small subset of the full category set.

This practical concession has two side-effects. 
The first pertains to the supertagger's inability to capture rare syntactic phenomena.
Although the percentage of sentences that may not be correctly analyzed due to the missing categories is usually relatively small, it still places an upper bound on the resulting parser's strength which is hard to ignore.
The second, and perhaps more far reaching, consequence is the implicit constraint it places on the grammar itself. 
Essentially, the grammar must be sufficiently coarse while also allocating most of its probability mass on a small number of unique categories. 
Grammars enjoying a higher level of analytical sophistication are practically unusable, since the associated supertagger would require prohibitive amounts of data to overcome their inherent sparsity.

We take a different view on the problem, instead treating it as sequence transduction. 
We propose a novel supertagger based on the Transformer architecture~\cite{vaswani2017attention} that is capable of constructing categories inductively, bypassing the aforementioned limitations.
We test our model on a highly-refined, automatically extracted type-logical grammar for written Dutch, where it achieves competitive results for high frequency categories, while acquiring the ability to treat rare and even unseen categories adequately.

\section{Type-Logical Grammars}

The type-logical strand of categorial grammar adopts a proof-theoretic perspective on natural language syntax and semantics: checking whether a phrase is syntactically well-formed amounts to a process of logical deduction deriving its type from the types of its constituent parts~\cite{book}.
What counts as a valid deduction depends on the type logic used. The type logic we aim for
is a variation on the simply typed fragment of Multiplicative Intuitionistic Linear Logic (MILL), where the type-forming operation of interest is linear implication (for a brief but instructive introduction, refer to~\citet{wadler1993taste}). Types are inductively defined by the
following grammar:
\begin{align}
\textsc{t} ::= \ & \textsc{a} \ | \  \textsc{t}_1 \myrightarrow{\text{d}} \textsc{t}_2
\label{eqn:induction}
\end{align}
where $\textsc{t}$, $\textsc{t}_1$, $\textsc{t}_2$ are types, $\textsc{a}$ is an atomic type and $\myrightarrow{d}$ an implication arrow, further subcategorized by the label $d$.

Atomic types are assigned to phrases that are considered `complete', e.g.~$\textsc{np}$ for noun phrase, $\textsc{pron}$ for pronoun, etc. Complex types, on the other hand, are the type signatures of binary functors that compose with a single word or phrase to produce a larger phrase; for instance $\textsc{np} \myrightarrow{\text{su}} \textsc{s}$ corresponds to a functor that consumes a noun phrase playing the subject role to create a sentence --- an intransitive verb.

The logic provides \emph{judgements} of the form $\Gamma\vdash B$, stating that from a multiset of assumptions $\Gamma=A_1,\ldots A_n$ one can derive conclusion $B$. In addition to the axiom $A\vdash A$, there are two rules of inference; implication elimination~(\ref{eqn:el}) and implication introduction~(\ref{eqn:int})\footnote{For labeled implications $\myrightarrow{d}$, we make sure that composition is with respect to the $d$ dependency relation.}.
Intuitively, the first says that if one has a judgement of the form $\Gamma \vdash A \rightarrow B$ and a judgement of the form $\Delta \vdash A$, one can deduce that assumptions $\Gamma$ and $\Delta$ together derive a proposition $B$.
Similarly, the second says that if one can derive $B$ from assumptions $A$ and $\Gamma$ together, then from $\Gamma$ alone one can derive an implication $A\rightarrow B$.

\begin{equation}
\infer[\rightarrow E]{\Gamma, \Delta \vdash {B}}{
    \Gamma \vdash A \rightarrow B
    &
    \Delta \vdash A
    }
\label{eqn:el}
\end{equation}

\begin{equation}
\infer[\rightarrow I]{\Gamma \vdash A\rightarrow B}{
    A,\Gamma \vdash B
    }
\label{eqn:int}
\end{equation}

The view of language as a linear type system offers many meaningful insights. 
In addition to the mentioned correspondence between parse and proof, the Curry-Howard `proofs-as-programs' interpretation guarantees a direct translation
from proofs to computations.
The two rules necessary for proof construction have their computational analogues in function application and abstraction respectively, a link that paves the way to seamlessly move from a syntactic derivation to a program that computes the associated meaning in a compositional manner.

\section{Constructive Supertagging}
Categorial grammars assign denotational semantics to types, which are in turn defined via a set of inductive rules, as in~(\ref{eqn:induction}).
These, in effect, are the productions a simple, context-free grammar; a {\it grammar of types} underlying the grammar of sentences.
In this light, any type may be viewed as a word of this simple type grammar's language; a regularity which we can try to exploit.

Considering neural networks' established ability of implicitly learning context-free grammars~\cite{noPhysics}, it is reasonable to expect that, given enough representational capacity and a robust training process, a network should be able to learn a context-free grammar embedded within a wider sequence labeling task.
Jointly acquiring the two amounts to learning a) how to produce types, including novel ones, and b) which types to produce under different contexts, essentially providing all of the necessary building blocks for a supertagger with unrestricted co-domain.
To that end, we may represent a single type as a sequence of characters over a fixed vocabulary, defined as the union of atomic types and type forming operators (in the case of type-logical grammars, the latter being $n$-ary logical connectives).
A sequence of types is then simply the concatenation of their corresponding representations, where type boundaries can be marked by a special separation symbol.

The problem then boils down to learning how to transduce a sequence of words onto a sequence of unfolded types. 
This can be pictured as a case of sequence-to-sequence translation, operating on word level input and producing character level output, with the source language now being the natural language and the target language being the language defined by the syntax and semantics of our categorial grammar.

\section{Related Work}
Supertagging has been standard practice for lexicalized grammars with complex lexical entries since the work of \citet{supertagging}.
In its original formulation, the categorial assignment process is enacted by an N-gram Markov model.
Later work utilized Maximum Entropy models that account for word windows of fixed length, while incorporating expanded lexical features and POS tags as inputs~\cite{cc}.
During the last half of the decade, the advent of word embeddings caused a natural shift towards neural architectures, with recurrent neural networks being established as the prime components of recent supertagging models.
\citet{xu-etal-2015-ccg} first used simple RNNs for CCG supertagging, which were gradually succeeded by LSTMs~\cite{vaswani2016lstm, lewis-etal-2016-lstm}, also in the context of Tree-Adjoining Grammars~\cite{tag}.

Regardless of the particular implementation, the above works all fall in the same category of sequence labeling architectures.
As such, the type vocabulary (i.e. the set of candidate categories) is always considered fixed and pre-specified --- it is, in fact, hard coded within the architecture itself (e.g. in the network's final classification layer).
The inability of such systems to account for unseen types or even consistently predict rare ones has permeated through the training and evaluation process; a frequency cut-off is usually applied on the corpus, keeping only categories that appear at least 10 times throughout the training set~\cite{cc}.
This limitation has been acknowledged in the past; in the case of CCG, certain classes of syntactic constructions pose significant difficulties for parsing due to categories completely missing from the corpus~\cite{clark2004object}.
An attempt to address the issue was made in the form of an inference algorithm, which iteratively expands upon the lexicon with new categories for unseen words~\cite{expansion} --- its applicability, however, is narrow, as new categories can often be necessary even for words that have been previously encountered.

We differentiate from relevant work in not employing a type lexicon at all, fixed or adaptive.
Rather than providing our system with a vocabulary of types, we seek to instead encode the type construction process directly within the network.
Type prediction is no longer a discernible part of the architecture, but rather manifested via the network's weights as a dynamic generation process, much like a language model for types that is conditioned on the input sentence.

\section{Data}
\subsection{Corpus}
\begin{figure*}[t]
\begin{subfigure}{1\textwidth}
\scriptsize
\[
\infer[\rightarrow E]{\text{we} (\textit{we}), \text{geven} (\textit{give}), \text{enkele} (\textit{some}), \text{voorbeelden} (\textit{examples}) \vdash \textsc{s}_\text{main}}{
	\infer[L]{\text{we} \vdash \textsc{pron}}{}
	&
	\infer[\rightarrow E]{\text{geven}, \text{enkele}, \text{voorbeelden} \vdash \textsc{pron} \myrightarrow{su} \textsc{s}_\text{main}}{
		\infer[L]{\text{geven} \vdash \textsc{np} \myrightarrow{obj} \textsc{pron} \myrightarrow{su}\textsc{s}_\text{main}}{}
		&
		\infer[\rightarrow E]{\text{enkele}, \text{voorbeelden} \vdash \textsc{np}}{
			\infer[L]{\text{enkele} \vdash \textsc{np} \myrightarrow{det} \textsc{np}}{}
			&
			\infer[L]{\text{voorbeelden} \vdash \textsc{np}}{}
		}
	}
}
\]
\label{fig:voorbeelden}
\caption{Derivation for ``we geven enkele voorbeelden'' (\textit{we give some examples)}, showcasing a simple transitive verb derivation.}
\end{subfigure}
\begin{subfigure}{1\textwidth}
\scriptsize
\[
\infer[\rightarrow E]{\text{welke} (\textit{which}), \text{rol} (\textit{role}), \text{spelen} (\textit{play}), \text{typen} (\textit{types}) \vdash \textsc{whq}}{
	\infer[\rightarrow E]{\text{welke}, \text{rol} \vdash (\textsc{n} \myrightarrow{obj} \textsc{sv1}) \myrightarrow{body} \textsc{whq}}{
		\infer[L]{\text{welke} \vdash \textsc{n} \myrightarrow{det} (\textsc{n} \myrightarrow{obj} \textsc{sv1}) \myrightarrow{body} \textsc{whq}}{}
		&
		\infer[L]{\text{rol} \vdash \textsc{n}}{}
		}
	&
	\infer[\rightarrow I]{\text{spelen}, \text{typen} \vdash \textsc{n} \myrightarrow{obj} \textsc{sv1}}{
	\infer[\rightarrow E]{\text{spelen}, \text{typen}, \textsc{n} \vdash \textsc{sv1}}{
		\infer[L]{\text{typen} \vdash \textsc{np}}{}
		&
		\infer[\rightarrow E]{\text{spelen}, \textsc{n} \vdash \textsc{np} \myrightarrow{su} \textsc{sv1}}{	
			\infer[id]{\textsc{n} \vdash \textsc{n}}{}
			&
			\infer[L]{\text{spelen} \vdash \textsc{n} \myrightarrow{obj} \textsc{np} \myrightarrow{su} \textsc{sv1}}{}
		}	
	}
	}
	}
\]
\label{fig:role}
\caption{Derivation for ``welke rol spelen typen'' (\textit{which role do types play)}, showcasing object-relativisation via second-order types. Type \textsc{sv1} stands for verb-initial sentence clause.}
\end{subfigure}
\begin{subfigure}{1\textwidth}
\scriptsize
\[
\infer[\rightarrow E]{\text{een (\textit{a})}, \text{eenvoudig (\textit{simple})}, \text{en (\textit{and})}, \text{degelijk (\textit{solid})}, \text{idee (\textit{idea})} \vdash \textsc{np}}{
	\infer[L]{\text{een} \vdash \textsc{n} \myrightarrow{det} \textsc{np}}{}
	&
	\infer[\rightarrow E]{\text{eenvoudig}, \text{en}, \text{degelijk}, \text{idee} \vdash \textsc{n}}{
		\infer[]{\text{eenvoudig}, \text{en}, \text{degelijk} \vdash \textsc{n} \myrightarrow{mod} \textsc{n}}{
			\infer[L]{\text{eenvoudig} \vdash \textsc{adj}}{}
			&
			\infer[L]{\text{en} \vdash  \textsc{adj}^\ast \myrightarrow{cnj} \textsc{n} \myrightarrow{mod} \textsc{n}}{}
			&
			\infer[L]{\text{degelijk} \vdash \textsc{adj}}{}
		}
		&
		\infer[L]{\text{idee} \vdash \textsc{n}}{}
	}
}
\]
\label{fig:idea}
\caption{Derivation for ``een eenvoudig en degelijk idee'' (\textit{a simple and solid idea}), showcasing non-polymorphic conjunction of two adjectives forming a noun-phrase modifier.}
\end{subfigure}
\caption{Syntactic derivations of example phrases using our extracted grammar. Lexical type assignments are the proofs' axiom leaves marked $L$. Identity for non-lexically grounded axioms is marked $id$. Parentheses are right implicit. Phrasal heads are associated with complex (functor) types. Phrases are composed via function application of functors to their arguments (i.e. implication elimination: $\rightarrow E$). Hypothetical reasoning for gaps is accomplished via function abstraction of higher-order types (i.e. implication introduction: $\rightarrow I$).}
\label{fig:proofs}
\end{figure*}

The experiments reported on focus on Dutch, a language with relatively free word order
that allows us to highlight the benefits of our non-directional type logic.
For our data needs, we utilize the Lassy-Small corpus~\cite{Lassy}. 
Lassy-Small contains approximately 65000 annotated sentences of written Dutch, comprised of over 1 million words in total. 
The annotations are DAGs with syntactic category labels at the nodes, and
dependency labels at the edges. 
The possibility of re-entrancy obviates the need for abstract syntactic elements (gaps, traces, etc.) in the annotation of unbounded dependencies and related phenomena.

\subsection{Extracted Grammar}
To obtain type assignments from the annotation graphs, we design and apply an adaptation of Moortgat and Moot's~\shortcite{CGN} extraction algorithm.
Following established practice, we assign phrasal heads a functor (complex) type selecting for its dependents. 
Atomic types are instantiated by a translation table that maps part-of-speech tags and phrasal categories onto their corresponding types. 

As remarked above, we diverge from standard categorial practice by making no distinction between rightward and leftward implication (slash and backslash, respectively), rather collapsing both into the direction-agnostic linear implication. 
We compensate for the possible loss in word-order sensitivity by subcategorizing the implication arrow into a set of distinct linear functions, the names of which are instantiated by the inventory of dependency labels present in the corpus.
This decoration amounts to including the labeled dependency materialized by each head (in the context of a particular phrase) within its corresponding type, vastly increasing its informational content.
In practical terms, dependency labeling is no longer treated as a task to be solved by the downstream parser; it is now internal to the grammar's type system. 
To consistently binarize all of our functor types, we impose an obliqueness ordering~\cite{dowty} over dependency roles, capturing the degree of coherence between a dependent and the head.
 Figure~\ref{fig:proofs} presents a few example derivations, indicating how our grammar treats a selection of interesting linguistic phenomena.

\begin{figure}[h]
\hspace{-35pt}
\includegraphics[scale=0.21]{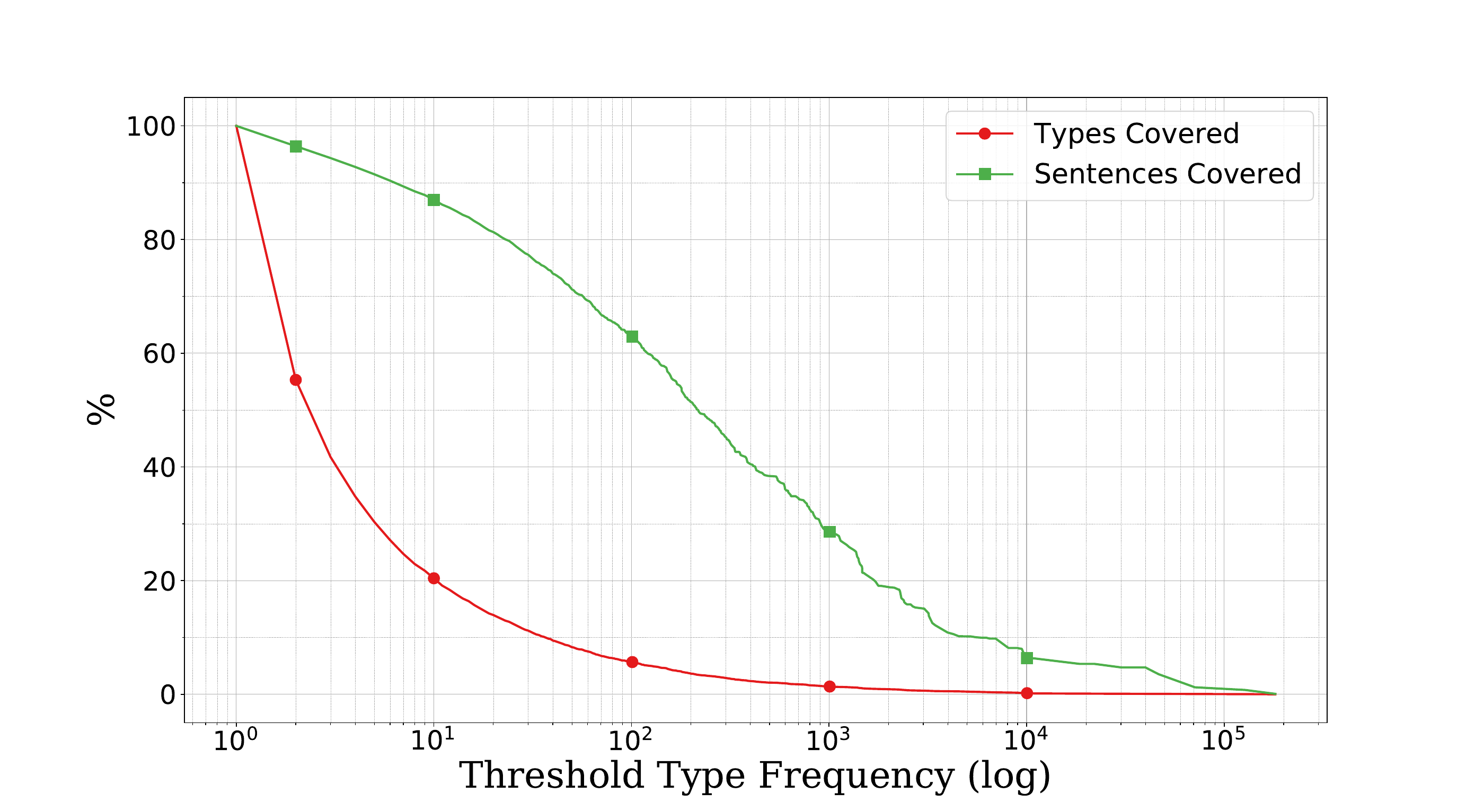}
\caption{Percentage of types and sentences covered as a function of type frequency. The vast majority of types (80\%) are rare (have less than 10 occurrences). At least one such type is present in a non-negligible part of the corpus (12\% of the overall sentences). A significant portion of types (45\%) appears just once throughout the corpus.}
\label{fig:stats}
\end{figure}

The algorithm's yield is a type-logical treebank, associating a type sequence to each sentence. 
The treebank counts approximately 5700 unique types, made out of 22 binary connectives (one for each dependency label) and 30 atomic types (one for each part-of-speech tag or phrasal category). 
As Figure~\ref{fig:stats} suggests, the comprehensiveness of such a fine-grained grammar comes at the cost of a sparser lexicon. 
Under this regime, recognizing rare types as first-class citizens becomes imperative.

Finally, given that all our connectives are of a fixed arity, we may represent types unambiguously using polish notation~\cite{hamblin1962translation}. 
Polish notation eliminates the need for brackets,  reducing the representation's length and succinctly encoding a type's arity in an up-front manner. 

\section{Model}
Even though prior work suggests that both the supertagging and the CFG-generation problems are learnable (at least to an extent) in isolation, the composition of the two is less straightforward.
Predicting the next atomic symbol requires for the network to be able to model local, close-range dependencies as ordained by the type-level syntax.
At the same time, it needs a global receptive field in order to correctly infer full types from distant contexts, in accordance with the sentence-level syntax.

Given these two requirements, we choose to employ a variant of the Transformer for the task at hand~\cite{vaswani2017attention}. 
Transformers were originally proposed for machine translation; treating syntactic analysis as a translation task is not, however, a new idea~\cite{foreign}. 
Transformers do away with recurrent architectures, relying only on self-attention instead, and their proven performance testifies to their strength.
Self-attention grants networks the ability to selectively shift their focus over their own representations of non-contiguous elements within long sequences, based on the current context, exactly fitting the specifications of our problem formulation.

Empirical evidence points to added benefits from utilizing language models at either side of an encoder-decoder architecture~\cite{D17-1039}. Adhering to this, we employ a pretrained Dutch ELMo~\cite{N18-1202,dutch_elmo} as large part of our encoder.

\subsection{Network}
\begin{figure*}[h]
\centering
\begin{tikzpicture}[every text node part/.style={align=center},
 every node/.style={transform shape},
 scale=0.7,
block/.style={rectangle, inner sep=0pt, minimum width=120pt, minimum height=60pt, rounded corners, ultra thick},
str/.style={rectangle, inner sep=0pt, minimum width=120pt, minimum height=20pt},
arrow/.style={->, ultra thick},
pwise/.style={circle, inner sep=0pt, minimum size=10pt},
smallblock/.style={circle, inner sep=5pt, minimum size=12pt, rounded corners, thick}]

	\node[str] (sentence) at (0, 5) {Input Sentence};		
	\node[str] (symbols) at (7.5, 5) {Output Sequence};

	\node[block, draw=black, fill=gray!10, draw=gray!130] (elmo) at (0,8) {\textcolor{gray!110}{ELMo}};
	\node[block, draw=black, fill=enc] (te) at (0,12) {\textbf{Encoder}};
	\node[block, draw=black, fill=emb] (se) at (7.5,8) {\textbf{Embedding}};
	\node[block, draw=black, fill=dec2] (td2) at (7.75, 12.25) {};
	\node[block, draw=black, fill=dec] (td) at (7.5,12) {\textbf{Decoder}};
	\node[block, draw=black, fill=emb] (set) at (15,12) {\textbf{Embedding}\\ (transposed)};	
	\node[smallblock, draw=black] (ss) at (15, 8) {$\sigma$};
	\node[smallblock, draw=black] (am) at (12, 8) {$\alpha$};
	\node[str] (out) at (15,5) {Output Probabilities};

	\draw (symbols) edge [arrow, gray!130] node[right] {M symbols} (se);
	\draw  (sentence) edge [arrow, gray!130] node[left] {N words} (elmo);
	\draw  (elmo) edge [arrow, gray!130] node[left] {Sentence Embedding\\ $\mathbb{R} ^ {N \times 1024}$} (te);
	\draw  (se) edge [arrow] node[right] {Symbol Embeddings\\ $\mathbb{R} ^ {M \times 1024}$} (td);
	\draw ($(te.east) + (0, 0.5)$) edge [arrow] node[above] {Encoder Keys\\ $\mathbb{R}^ {N \times 1024}$} ($(td.west) + (0, 0.5)$);
	\draw ($(te.east) + (0, -0.5)$) edge [arrow] node[below] {Encoder Values\\ $\mathbb{R}^ {N \times 1024}$} ($(td.west) + (0, -0.5)$);\
	\draw ($(td.east) + (0.25, 0)$) edge [arrow] node[above] {Decoder Values\\ $\mathbb{R}^ {M \times 1024}$} (set);
	\draw (set) edge [arrow] node[right] {Class Weights} (ss);
	\draw (ss) edge [arrow] (out);
	\draw (set.south) [dotted, very thick] .. controls +(-1,0) and +(-0.5, 0.5) .. (am.north);
	\draw (am.south) [dotted, very thick, ->] .. controls +(0,-0.5) and +(2, ) .. (symbols.east);
\end{tikzpicture}
\caption{The model architecture, where $\sigma$ and $\alpha$ denote the \textit{sigsoftmax} and \textit{argmax} functions respectively, grayed out components indicate non-trainable components and the dotted line depicts the information flow during inference.}
\label{fig:net}
\end{figure*}
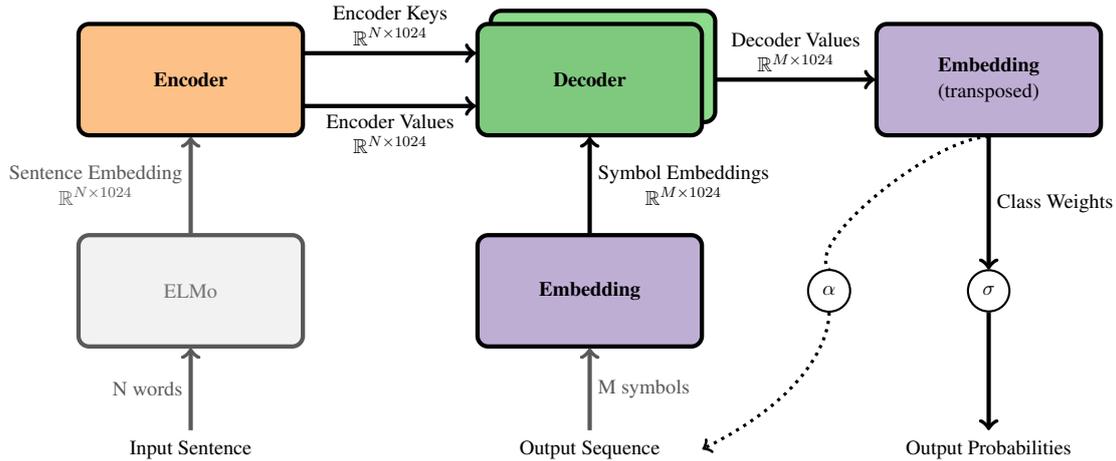

\begin{figure*}[h]
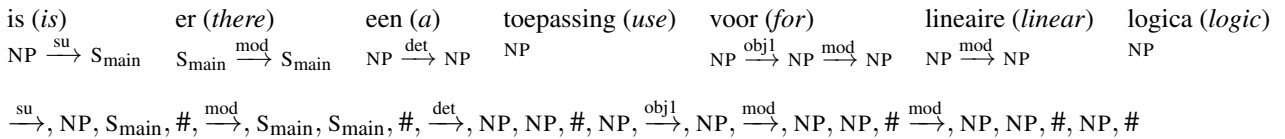

\centering
\begin{minipage}{1.2\textwidth}
\gll {is (\textit{is})} \quad {er (\textit{there})} \quad {een (\textit{a})} \quad {toepassing (\textit{use})} \quad {voor (\textit{for})} \quad {lineaire (\textit{linear})} \quad {logica (\textit{logic})}\\
$\textsc{np}\myrightarrow{su}\textsc{s}_\text{main}$
\quad
$\textsc{s}_\text{main}\myrightarrow{mod}\textsc{s}_\text{main}$
\quad
\small $\textsc{np}\myrightarrow{det}\textsc{np}$
\quad
\small $\textsc{np}$
\quad
\small $\textsc{np}\myrightarrow{obj1}\textsc{np}\myrightarrow{mod}\textsc{np}$
\quad
\small $\textsc{np}\myrightarrow{mod}\textsc{np}$
\quad
\small $\textsc{np}$\\
\trans $\myrightarrow{su},\textsc{np}, \textsc{s}_\text{main}, \text{\#}, \myrightarrow{mod},\textsc{s}_\text{main}, \textsc{s}_\text{main}, \text{\#},\myrightarrow{det},\textsc{np}, \textsc{np}, \text{\#},\textsc{np},\myrightarrow{obj1},\textsc{np}, \myrightarrow{mod},\textsc{np}, \textsc{np}, \text{\#}\myrightarrow{mod},\textsc{np}, \textsc{np}, \text{\#},\textsc{np},\text{\#}$
\end{minipage}
\caption{Input-output example pair for the sentence ``is er een toepassing voor lineaire logica?'' (\textit{is there a use for linear logic?}). The first two lines present the input sentence and the types that need to be assigned to each word. The third line presents the desired output sequence, with types decomposed to atomic symbol sequences under polish notation, and \# used as a type separator.}
\label{fig:io}
\end{figure*}

Our network follows the standard encoder-decoder paradigm. A high-level overview of the architecture may be seen in Figure~\ref{fig:net}. The network accepts a sequence of words as input, and as output produces a (longer) sequence of tokens, where each token can be an atomic type, a logical connective or an auxiliary separation symbol that marks type boundaries.
An example input/output pair may be seen in Figure~\ref{fig:io}.

Our encoder consists of a frozen ELMo followed by a single Transformer encoder layer. 
The employed ELMo was trained as a language model and constructs contextualized, 1024-dimensional word vectors, shown to significantly benefit downstream parsing tasks.
To account for domain adaptation without unfreezing the over-parameterized ELMo, we allow for a transformer encoder layer of 3 attention heads to process ELMo's output\footnote{Given that no gradient flow is allowed past the transformer encoder layer, in practice we compute the ELMo embeddings of our input sentences in advance, and feed those onto the rest of the network.}.

Our decoder is a 2-layer Transformer decoder.
Since the decoder processes information at a different granularity scale compared to the encoder, we break the usual symmetry by setting its number of attention heads to 8.

At timestep $t$, the network is tasked with modeling the probability distribution of the next atomic symbol $a_t$, conditional on all previous predictions $a_0,\ \dots, a_{t-1}$ and the whole input sentence $w_0,\ \dots, w_\tau$, and parameterized by its trainable weights $\theta$: 
\vspace{-0.5em}
$$p_\theta(a_t | a_0,\ \dots,\ a_{t-1},\ w_0,\ \dots, w_\tau)$$

We make a few crucial alterations to the original Transformer formulation. 

First, for the separable token transformations we use a two-layer, dimensionality preserving, feed-forward network.
We replace the rectifier activation of the intermediate layer with the empirically superior Gaussian Error Linear Unit~\cite{gelu}.

Secondly, since there are no pretrained embeddings for the output tokens, we jointly train the Transformer alongside an atomic symbol embedding layer.
To make maximal use of the extra parameters, we use the transpose of the embedding matrix to convert the decoder's high-dimensional output back into token class weights.
We obtain the final output probability distributions by applying sigsoftmax~\cite{kanai2018sigsoftmax} on these weights.

\subsection{Training}
We train our network using the adaptive training scheme proposed by Vaswani et al~\shortcite{vaswani2017attention}.
We apply stricter regularization by increasing both the dropout rate and the redistributed probability mass of the Kullback-Leibler divergence loss to $0.2$.
The last part is of major importance, as it effectively discourages the network from simply memoizing common type patterns.

\section{Experiments and Results}
In all described experiments, we run the model\footnote{The code for the model and processing scripts can be found at \url{https://github.com/konstantinosKokos/Lassy-TLG-Supertagging}.} on the subset of sample sentences that are at most 20 words long. 
We use a train/val/test split of 80/10/10\footnote{It is worth pointing out that the training set contains only $\sim$85\% of the overall unique types, the remainder being present only in the validation and/or test sets.}.
We train with a batch size of 128, and pad sentences to the maximum in-batch length. 
Training to convergence takes, on average, eight hours \& 300 epochs for our training set of 45000 sentences on a GTX1080Ti. 
We report averages over 5 runs. 

Accuracy is reported on the type-level; that is, during evaluation, we predict atomic symbol sequences, then collapse subtype sequences into full types and compare the result against the ground truth. 
Notably, a single mistake within a type is counted as a completely wrong type. 

\subsection{Main Results}
We are interested in exploring the architecture's potential at supertagging, as traditionally formulated, as well as its capacity to learn the grammar beyond the scope of the types seen in the training data.
We would like to know whether the latter is at all possible (and, if so, to what degree), but also whether switching to a constructive setting has an impact on overall accuracy.

\paragraph{Digram Encoding}
Predicting type sequences one atomic symbol or connective at a time provides the vocabulary to construct new types, but results in elongated target output sequence lengths\footnote{Note that if lexical categories are, on average, made out of $c$ atomic symbols, the overall output length is a constant factor of the sentence length, i.e. there is no change of complexity class with respect to a traditional supertagger.}.
As a countermeasure, we experiment with {\it digram encoding}, creating new atomic symbols by iteratively applying pairwise merges of  the most frequent intra-type symbol digrams~\cite{bpe}, a practice already shown to improve generalization for translation tasks~\cite{bpe2}. 
To evaluate performance, we revert the merges back into their atoms after obtaining the predictions.

With no merges, the model has to construct types and type sequences using only atomic types and connectives.
As more merges are applied, the model gains access to extra short-hands for subsequences within longer types, reducing the target output length, and thus the number of interactions it has to capture.
This, however, comes at the cost of a reduced number of full-type constructions effectively seen during training, while also increasing the number of implicit rules of the type-forming context-free grammar.
If merging is performed to exhaustion, all types are compressed into single symbols corresponding to the indivisible lexical types present in the treebank. 
The model then reduces to a traditional supertagger, never having been exposed to the internal type syntax, and loses the potential to generate new types.

We experiment with a fully constructive model employing no merges ($\text{M}_0$), a fully merged one i.e. a traditional supertagger, ($\text{M}_\infty$), and three in-between models trained with 50, 100 and 200 merges ($\text{M}_{50}$, $\text{M}_{100}$ and $\text{M}_{200}$ respectively).
Table~\ref{table:numbers} displays the models' accuracy. 
In addition to the overall accuracy, we show accuracy over different bins of type frequencies, as measured in the training data: unseen, rare (1-10), medium (10-100) and high-frequency ($>$ 100) types.

\begin{table}[h]
\noindent
\newcommand{\ra}[1]{\renewcommand{\arraystretch}{#1}}
\ra{1.1}
\hspace{-10pt}
\begin{tabularx}{0.49\textwidth}{@{}Xsssss@{}}
&  \multicolumn{5}{c}{\centering Type Accuracy}\\
\cmidrule{2-6}
\multicolumn{1}{l}{}  &  \small Overall & \small Unseen & \small  Freq & \small Freq & \small Freq \\
\multicolumn{1}{l}{Model}  &  \small  & \small Types & \small  1-10 & \small 10-100 & \small $>$100 \\

\toprule[0.01em]
\centering $\text{M}_{0}$  & \textbf{ 88.05} & \textbf{19.2} & \textbf{45.68} & \textbf{65.62} & 89.93\\
\midrule[0.001em]
\centering $\text{M}_{50}$  & 88.03 & 15.97 & 43.69 & 64.33 & \textbf{90.01}\\
\centering $\text{M}_{100}$ & 87.87 & 15.02 & 41.61 & 63.71 & 89.9 \\
\centering $\text{M}_{200}$ & 87.54 & 11.7 & 39.56 & 62.4 & 89.64\\
\midrule[0.001em]
\centering $\text{M}_{\infty}$ & 87.2 & - & 23.91 & 59.03 & 89.89\\
\end{tabularx}
\caption{Model performance at different merge scales, with respect to training set type frequencies. $\text{M}_i$ denotes the model at $i$ merges, where $\text{M}_\infty$ means the fully merged model. For the fully merged model there is a 1 to 1 correspondence between input words and output types, so we do away with the separation symbol.}
\label{table:numbers}
\end{table}

Table 1 shows that all constructive models perform overall better than $\text{M}_{\infty}$, owing to a consistent increase in their accuracy over unseen, rare, and mid-frequency types.
This suggests significant benefits to using a representation that is aware of the type syntax.
Additionally, the gains are greater the more transparent the view of the type syntax is, i.e. the fewer the merges.
The merge-free model $\text{M}_0$ outperforms all other constructive models across all but the most frequent type bins, reaching an overall accuracy of 88.05\% and an unseen category accuracy of 19.2\%.

We are also interested in quantifying the models' ``imaginative'' precision, i.e.,  how often do they generate new types to analyze a given input sentence, and, when they do, how often are they right (Table~\ref{table:imagination}). 
Although all constructive models are eager to produce types never seen during training, they do so to a reasonable extent. 
Similar to their accuracy, an upwards trend is also seen in their precision, with $\text{M}_0$ getting the largest percentage of generated types correct. 

Together, our results indicate that the type-syntax is not only learnable, but also a representational resource that can be utilized to tangibly improve a supertagger's generalization and overall performance.

\begin{table}[t]
\centering
\noindent
\newcommand{\ra}[1]{\renewcommand{\arraystretch}{#1}}
\ra{1.1}
\begin{tabularx}{0.49\textwidth}{@{}Xsss@{}}
\multicolumn{1}{l}{Model}  &  New Types & Unique & Correct \small (\%)\\
\multicolumn{1}{l}{}  &  Generated  &  &  \\
\toprule[0.01em]
\centering $\text{M}_{0}$  & 213.6 & 199.2 &  \textbf{44.39} \small (\textbf{20.88}) \\
\centering $\text{M}_{50}$  & 186.6 & 174.2 & 37.89 \small (20.3) \\
\centering $\text{M}_{100}$  & 187.8 & 173.4 & 34.31 \small (18.27) \\
\centering $\text{M}_{200}$  & 190.4 & 178.8 & 27.46 \small (14.42) \\
\end{tabularx}
\caption{Repetition-averaged unseen type generation and precision.}
\label{table:imagination}
\end{table}

\subsection{Other Models}
Our preliminary experiments involved RNN-based encoder-decoder architectures.
We first tried training a single-layer BiGRU encoder over the ELMo representations, connected to a single-layer GRU decoder, following~\citet{encdec}; the model took significantly longer to train and yielded far poorer results (less than 80\% overall accuracy and a strong tendency towards memoizing common types).
We hypothesize that the encoder's fixed length representation is unable to efficiently capture all of the information required for decoding a full sequence of atomic symbols, inhibiting learning.

As an alternative, we tried a separable LSTM decoder operating individually on the encoder's representations of each word. 
Even though this model was faster to train and performed marginally better compared to the previous attempt, it still showed no capacity for generalization over rarer types. 
This is unsurprising, as this approach assumes that the decoding task can be decomposed at the type-level; crucially, the separable decoder's prediction over a word cannot be informed by its predictions spanning other words, an information flow that evidently facilitates learning and generalization.

\section{Analysis}
\subsection{Type Syntax}
To assess the models' acquired grasp of the type syntax, we inspect type predictions in isolation.
Across all merge scales and consistently over all trained models, all produced types (including unseen ones) are \textit{well-formed}, i.e. they are indeed words of the type-forming grammar.
Further, the types constructed are fully complying with our implicit notational conventions such as the obliqueness hierarchy.

Even more interestingly, for models trained on non-zero merges it is often the case that a type is put together using the correct atomic elements that together constitute a merged symbol, rather than the merged shorthand trained on.
Judging from the above, it is apparent that the model gains a functionally complete understanding of the type-forming grammar's syntax, i.e. the means through which atomic symbols interact to produce types.

\subsection{Sentence Syntax}
\begin{figure*}
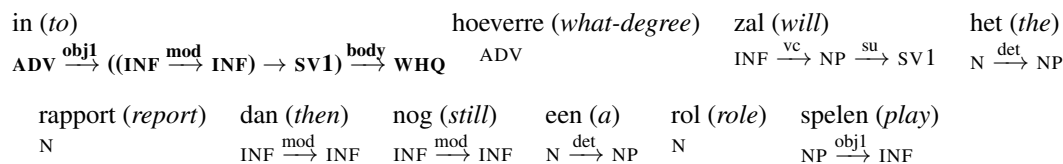

\centering
\begin{minipage}{1\textwidth}

\gll {in (\textit{to})}  {hoeverre (\textit{what-degree})} \quad {zal (\textit{will})} \quad {het (\textit{the})}\\
\small $\textsc{\textbf{adv}}\myrightarrow{\textbf{obj1}}\textbf{((}\textsc{\textbf{inf}}\myrightarrow{\textbf{mod}}\textsc{\textbf{inf}}\textbf{)}\rightarrow\textsc{\textbf{sv1}}\textbf{)}\myrightarrow{\textbf{body}}\textsc{\textbf{whq}}$
\quad 
\small $\textsc{adv}$
\quad
\small $\textsc{inf}\myrightarrow{vc}\textsc{np}\myrightarrow{su}\textsc{sv1}$
\quad
\small $\textsc{n}\myrightarrow{det}\textsc{np}$
\\

\gll \quad {rapport (\textit{report})} \quad {dan (\textit{then})} \quad {nog (\textit{still})} \quad {een (\textit{a})} \quad {rol (\textit{role})} \quad {spelen (\textit{play})}\\
\quad
\small $\textsc{n}$
\quad
\small $\textsc{inf}\myrightarrow{mod}\textsc{inf}$
\quad
\small $\textsc{inf}\myrightarrow{mod}\textsc{inf}$
\quad
\small $\textsc{n}\myrightarrow{det}\textsc{np}$
\quad
\small $\textsc{n}$
\quad
\small $\textsc{np}\myrightarrow{obj1}\textsc{inf}$\\

\end{minipage}
\caption{Type assignments for the correctly analyzed wh-question ``in hoeverre zal het rapport dan nog een rol spelen'' (\textit{to what extent will the report still play a role}) involving a particular instance of \textit{pied-piping}. The type of ``in'' was never seen during training; it consumes an adverb as its prepositional object, to then provide a third-order type that turns a verb-initial clause with a missing infinitive modifier into a wh-question. Such constructions are a common source of errors for supertaggers, as different instantiations require unique category assignments.}
\label{fig:electricity}
\end{figure*}

Beyond the spectrum of single types, we examine type assignments in context.

We first note a remarkable ability to correctly analyze syntactically complex constructions requiring higher-order reasoning, even in the presence of unseen types.
An example of such an analysis is shown in Fig~\ref{fig:electricity}.

For erroneous analyses, we observe a strong tendency towards self-consistency.
In cases where a type construction is wrong, types that interact with that type (as either arguments or functors) tend to also follow along with the mistake.
On one hand, this cascading behavior has the effect of increasing error rates as soon as a single error has been made.
On the other hand, however, this is a sign of an implicitly acquired notion of phrase-wide well-typedness, and exemplifies the learned long-range interdependencies between types through the decoder's auto-regressive formulation.
On a related note, we recognize the most frequent error type as misconstruction of conjunction schemes. 
This was, to a degree, expected, as coordinators display an extreme level of lexical ambiguity, owing to our extracted grammar's massive type vocabulary. 

\subsection{Output Embeddings}
Our network trains not only the encoder-decoder stacks, but also an embedding layer of atomic symbols.
We can extract this layer's outputs to generate vectorial representations of atomic types and binary connectives, which essentially are high-dimensional character-level embeddings of the type language.

Considering that dense supertag representations have been shown to benefit parsing~\cite{tag}, our atomic symbol embeddings may be further utilized by downstream tasks, as a highly refined source of type-level information.

\subsection{Comparison}
Our model's overall accuracy lies at 88\%, which is comparable to the state-of-the-art in TAG supertagging~\cite{tag} but substantially lower than CCG~\cite{clark2018semi}.
A direct numeric comparison holds little value, however, due to the different corpus, language and formalism used.
To begin with, our scores are the result of a more difficult problem, since our target grammar is far more refined. 
Concretely, we measure accuracy over a set 5700 types, which is one order of magnitude larger than the CCGBank test bed (425 in most published work; CCGBank itself contains a little over 1100 types) and 20\% larger than the set of TAGs in the Penn Treebank.
Practically, a portion of the error mass is allotted to mislabeling the implication arrow's name, which is in one-to-one correspondence with a dependency label of the associated parse tree.
In that sense, our error rate is already accounting for a portion of the labeled attachment score, a task usually deferred to a parser further down the processing line.
Further, the prevalence of entangled dependency structures in Dutch renders its syntax considerably more complicated than English.	

\section{Conclusion and Future Work}
Our paper makes three novel contributions to categorial grammar parsing. 
We have shown that attention-based frameworks, such as the Transformer, may act as capable and efficient supertaggers, eliminating the computational costs of recurrence. 
We have proposed a linear type system that internalizes dependency labels, expanding upon categorial grammar supertags and easing the burden of downstream parsing.
Finally, we have demonstrated that a subtle reformulation of the supertagging task can lift the closed world assumption, allowing for unbounded supertagging and stronger grammar learning while incurring only a minimal cost in computational complexity.

Hyper-parameter tuning and network optimization were not the priority of this work; it is entirely possible that different architectures or training algorithms might yield better results under the same, constructive paradigm.
This aside, our work raises three questions that we are curious to see answered. 
First and foremost, we are interested to examine how our approach performs under different datasets, be it different grammar specifications, formalisms or languages, as well as its potential under settings of lesser supervision.
A natural continuation is also to consider how our supertags and their variable-length, content-rich vectorial representations may best be integrated with a neural parser architecture.
Finally, given the close affinity between syntactic derivations, logical proofs and programs for meaning computation, we plan to investigate how insights on semantic compositionality may be gained from the vectorial representations of types and type-logical derivations.

\section*{Acknowledgments}
The authors would like to thank the anonymous reviewers for their remarks and suggestions.
The first two authors are supported by a NWO grant under the scope of the project ``A composition calculus for vector-based semantic modelling with a localization for Dutch'' (360-89-070).
The first author would like to acknowledge helpful feedback and suggestions from Richard Moot and Vasilis Bountris.

\bibliography{sources}
\bibliographystyle{acl_natbib}

\end{document}